\documentclass[letterpaper]{article} 
\usepackage{aaai23}  
\usepackage{times}  
\usepackage{helvet}  
\usepackage{courier}  
\usepackage[hyphens]{url}  
\usepackage{graphicx} 
\usepackage{natbib}  
\usepackage{caption} 
\usepackage{comment}
\usepackage{booktabs}
\usepackage{multirow}
\usepackage{bm}
\usepackage{mathrsfs}
\usepackage{amsthm,amsmath,amssymb}

\usepackage{algorithm}
\usepackage{algorithmic}

\usepackage{newfloat}
\usepackage{listings}
\DeclareCaptionStyle{ruled}{labelfont=normalfont,labelsep=colon,strut=off} 
\floatstyle{ruled}
\newfloat{listing}{tb}{lst}{}
\floatname{listing}{Listing}
\def\ie{\emph{i.e.}}
\def\eg{\emph{e.g.}}

\def\etc{{\em etc.}}

\title{MulGT: Multi-Task Graph-Transformer with Task-Aware Knowledge Injection and Domain Knowledge-Driven Pooling for Whole Slide Image Analysis}
\author{
    Weiqin Zhao\textsuperscript{\rm 1},
    Shujun Wang\textsuperscript{\rm 2},
    Maximus Yeung\textsuperscript{\rm 1},
    Tianye Niu\textsuperscript{\rm 3},
    Lequan Yu\textsuperscript{\rm 1,{\footnote{Corresponding author.}}}
}
\affiliations{

    \textsuperscript{\rm 1}  The University of Hong Kong, Hong Kong SAR, China\\
    \textsuperscript{\rm 2}  University of Cambridge, Cambridge, UK\\
    \textsuperscript{\rm 3}  Shenzhen Bay Laboratory, Shenzhen, China\\
    wqzhao98@connect.hku.hk, sw991@cam.ac.uk, mcfyeung@pathology.hku.hk, niuty@szbl.ac.cn, lqyu@hku.hk
}

\usepackage{bibentry}

\begin{document}

\maketitle
\begin{abstract}

%
Whole slide image (WSI) has been widely used to assist automated diagnosis under the deep learning fields.
%
%
However, most previous works only discuss the SINGLE task setting which is not aligned with real clinical setting, where pathologists often conduct multiple diagnosis tasks simultaneously.
%
Also, it is commonly recognized that the multi-task learning paradigm can improve learning efficiency by exploiting commonalities and differences across multiple tasks. 
To this end, we present a novel multi-task framework (\ie, MulGT) for WSI analysis by the specially designed Graph-Transformer equipped with Task-aware Knowledge Injection and Domain Knowledge-driven Graph Pooling modules.
%
Basically, with the Graph Neural Network and Transformer as the building commons, our framework is able to learn task-agnostic low-level local information as well as task-specific high-level global representation.
%
Considering that different tasks in WSI analysis depend on different features and properties, we also design a novel Task-aware Knowledge Injection module to transfer the task-shared graph embedding into task-specific feature spaces to learn more accurate representation for different tasks.
%
Further, we elaborately design a novel Domain Knowledge-driven Graph Pooling module for each task to improve both the accuracy and robustness of different tasks by leveraging different diagnosis patterns of multiple tasks.
%
We evaluated our method on two public WSI datasets from TCGA projects, \ie, esophageal carcinoma and kidney carcinoma.
Experimental results show that our method outperforms single-task counterparts and the state-of-the-art methods on both tumor typing and staging tasks.
%

\end{abstract}

\section{Introduction}

\begin{figure}[ht]
\centering
\includegraphics[width=1.0\linewidth]{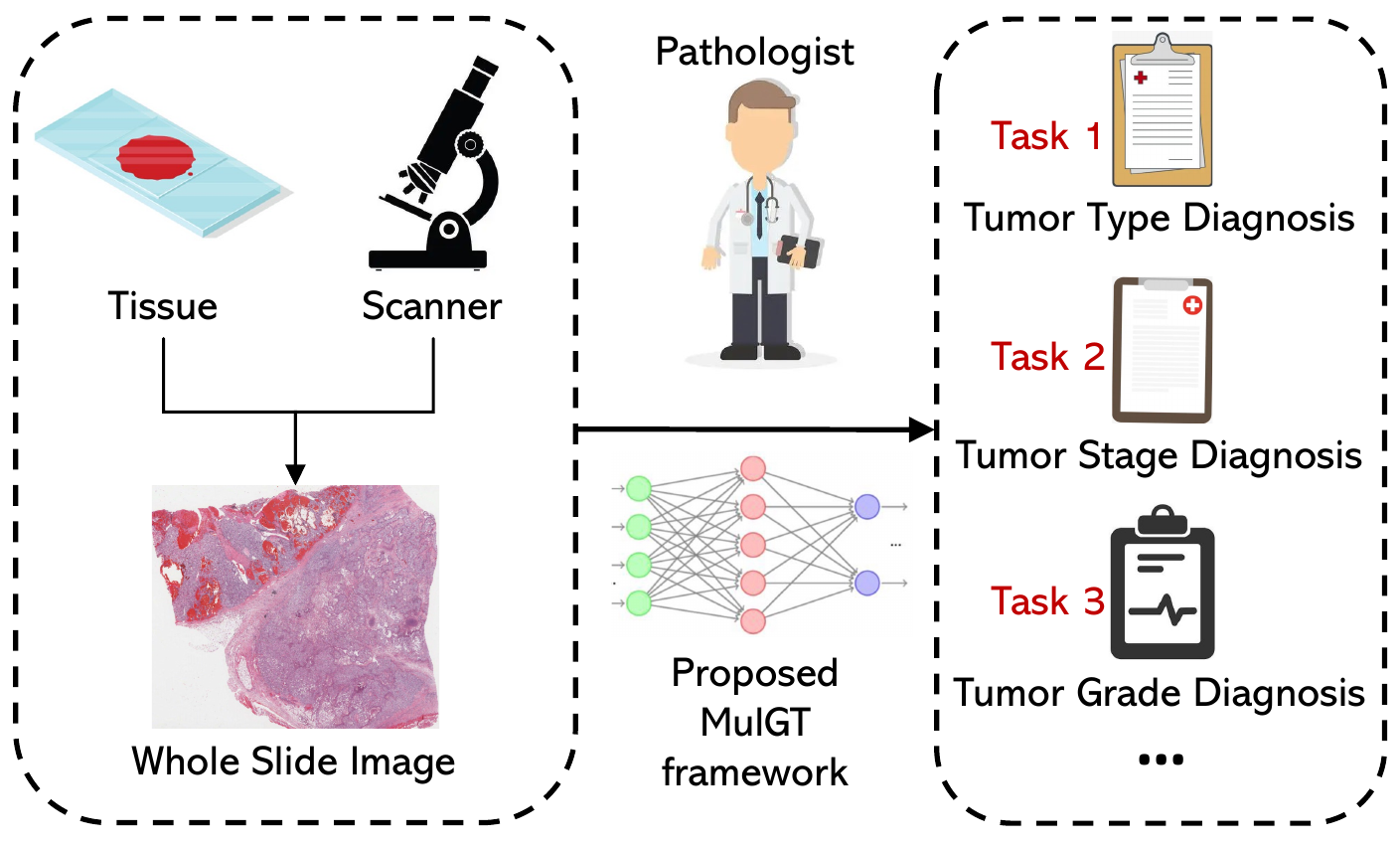}
\caption{Illustration of the multi-task learning setting for WSI analysis.} 
\label{Diagnosis_Process}
\vspace{-0.5cm}
\end{figure}

Histopathology analysis is the gold standard method for cancer diagnosis and prognosis.
Experienced pathologists can provide accurate analysis of biopsy specimens based on whole slide image (WSI), \ie, the high-resolution digitalization of the entire histology slide~\cite{khened2021generalized, pataki2022huncrc}. 
However, analyzing the WSIs is time-consuming and laborious due to the massive size of the WSIs and the complex colors and patterns of different tissue structures. 
To elevate the precision and speed of the examination, extensive research tools have been developed for automate computational WSI inspection~\cite{wang2019weakly, wang2019pathology, coudray2018classification}.

Due to the powerful expressivity of neural networks, many deep learning-based methods have been proposed for WSI analysis recently.
However, WSIs usually have a huge size (\eg, $150,000\times 150,000$), and it will be expensive to obtain detailed pixel-level annotations.
To overcome such challenges, multiple instance learning (MIL)~\cite{maron1998framework} becomes a promising direction to analyze WSI from slide-level annotations.
%
%
%
Specifically, MIL-based approaches first extract the feature embeddings of image tiles (\ie~patches) with a Convolution Network (CNN)~\cite{he2016deep, riasatian2021fine} or Vision Transformer Network (ViT)~\cite{chen2022scaling}.
%
Then, the feature embeddings are fed into an aggregation network to produce the slide-level predictions. 
Various network architectures have been employed to aggregate the information, including Graph Neural Network (GNN)~\cite{hou2022h2, guan2022node}, Transformer network~\cite{chen2022scaling, wang2022lymph}, and \etc~
Currently, Graph-Transformer architecture has also been introduced into WSI analysis~\cite{zheng2021deep} due to its nature to extract both low-level local features and high-level global information from the graphs (\ie WSIs).

However, most of the above works were limited to the setting with a single task, while pathologists often conducts more than one diagnosis results for one particular WSI (per patient), as shown in Figure~\ref{Diagnosis_Process}.
Besides, it is believed that multi-task learning paradigm
can improve learning efficiency and prediction accuracy by exploiting commonalities and differences across tasks.
Although there are also some works~\cite{yang2020detecting, vuong2020multi, murthy2017center} discussing multi-task learning in WSI analysis. 
They were designed for the patch-level prediction tasks instead of the slide-level ones. 
Therefore, they require patch-level annotations for training and hardly to be extended to the weakly-supervised slide-level label prediction directly. 
To address the above issues, we present a multi-task Graph-Transformer framework (\ie, MulGT) to conduct multiple slide-level diagnosis tasks simultaneously.

Our framework leverages the architecture of Graph-Transformer from two aspects: (1) learning task-agnostic low-level representation with a shared GNN, and (2) learning task-specific high-level representation with independent Transformer branches.
Meanwhile, considering that different tasks in WSI analysis usually require different features and properties of the tissue, we thereby design a novel Task-aware Knowledge Injection module in our framework to transfer the task-shared graph embedding into task-specific feature spaces via the cross-attention mechanism with a set of trainable task-specific latent tokens.
%
%
Furthermore, to reduce the computation cost, a graph pooling layer is usually adopted between the GNN part and the Transformer part in the Graph-Transformer architecture.
%
However, no attention has been paid to discussing the relationship among tasks or the graph pooling methods. 
In this paper, we are the first to argue that, in order to boost the performance of the Graph-Transformer architecture, it is necessary to design a task-aware pooling method to meet the different requirements of different downstream tasks. 
Especially in multi-task learning settings, the graph pooling methods should vary in different task branches if the nature of the tasks is different. 
%
Therefore, we elaborately design a novel Domain Knowledge-driven Graph Pooling module in our framework to improve both the accuracy and robustness of different task branches by leveraging the different diagnosis patterns of multiple WSI analysis tasks.

Our main contributions can be summarized as follows.
\begin{itemize}
    \item We devise a novel multi-task Graph-Transformer for slide-level WSI analysis. Different from methods, our framework conducts multiple diagnosis tasks simultaneously, thus benefiting from learning both the commonalities and differences of multiple tasks. 
    Extensive experiments with promising results on two public WSI datasets validate the effectiveness of our designed framework.
    
    \item To learn task-specific features, we design a novel Task-aware Knowledge Injection module to transfer the task-shared feature into task-specific feature spaces via the cross-attention mechanism with the latent tokens that contain task-specific knowledge.
    
    \item To import the prior knowledge from different diagnosis patterns of different tasks, we elaborately design a novel Domain Knowledge-driven Graph Pooling module to represent the information of the whole graph more properly for different tasks, facilitating the prediction process and reducing the computation cost.
    
\end{itemize}

\begin{figure*}[ht]
\centering
\includegraphics[width=0.9\textwidth]{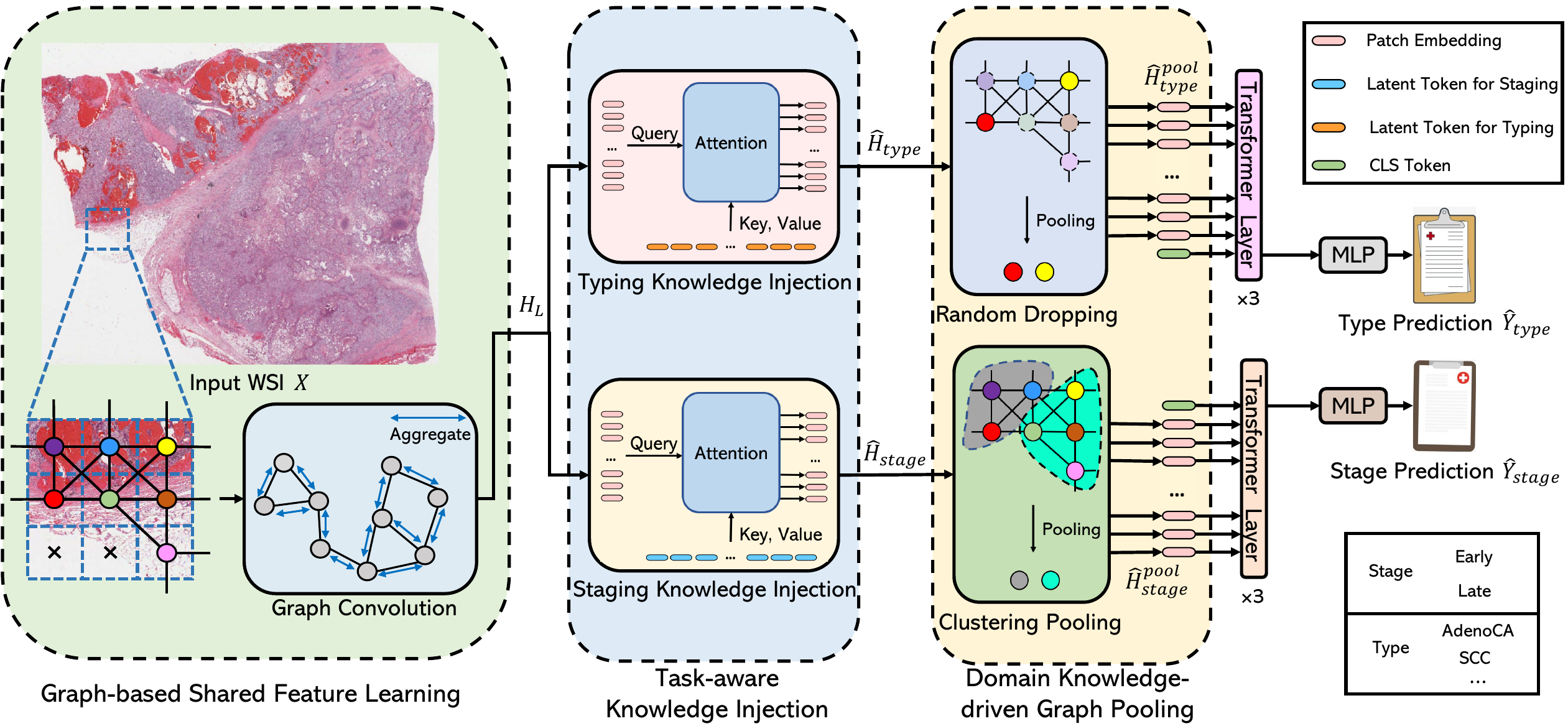}
\caption{Overview of the proposed MulGT framework. Patches are extracted from WSIs and abstracted as graph nodes. Follow the multi-task learning paradigm, the GNN part served as the task-shared layers to learn task-agnostic low-level local representation, while our proposed Task-aware Knowledge Injection and Domain Knowledge-drive Graph Pooling modules together with the transformer stack served as the task-independent layers to learn accurate high-level global representation.
} 
\label{GT_Framwork_Overview}
\end{figure*}
\section{Related Work}

\subsubsection{Multiple Instance Learning for WSI.}
Multiple instance learning (MIL) methods are widely used for WSI analysis and can be categorized into two paradigms: (1) instance-level methods and (2) embedding-level methods~\cite{amores2013multiple}. 
%
%
Generally, instance-level methods typically focus more on local information, while embedding-level methods emphasize global representation.
%
Several recent works adopted attention mechanisms into MIL for WSI analysis for instance aggregation. 
Particularly, the attention-based approach is able to identify the contribution of different instances during the global aggregation, like ABMIL~\cite{ilse2018attention}, DeepAttnMIL~\cite{yao2020whole}, and CLAM~\cite{lu2021data}. 
%
%
%
Recently, Graph-based and Transformer-based methods have also been utilized in computational pathology, as WSI instances could be abstracted as nodes of a graph or tokens of Transformer architecture. 
For example, H2Graph~\cite{hou2022h2} built a heterogeneous graph with different resolutions of WSI to learn a hierarchical representation, while HIPT~\cite{chen2022scaling} introduced a new ViT architecture to learn from the natural image hierarchical structure inherent in WSIs. 
However, most of the previous works were limited to the single task setting for slide-level analysis.

\subsubsection{Multi-task Learning.}

Multi-task learning~\cite{caruana1997multitask} jointly optimizes a set of tasks with hard or soft parameter sharing. 
It is well known that learning multiple tasks simultaneously can offer several advantages, including improved data efficiency and reduced overfitting through the regularization among multiple tasks~\cite{DBLP:journals/corr/abs-2009-09796}.
Some previous literature leveraged the relationship among multiple tasks in an explicit way. 
%
For example, ML-GCN~\cite{chen2019multi} built a directed graph over different object labels to facilitate multi-label image recognition, where each node is one particular object (\ie task) and edges are object correlations.
Meanwhile, some works~\cite{kendall2018multi, chen2018gradnorm} adopted adaptive weights for different tasks to balance the training process, while Liu \textit{et. al.}~\shortcite{liu2021conflict} introduced gradient-based methods to mitigate the negative transfer across tasks. 
Especially, RotoGrad~\cite{javaloy2021rotograd} used a set of rotation matrices to rotate the task-shared features into different feature spaces before the task-specific branches to avoid the gradient conflict among tasks. 
Partially inspired that transferring task-shared features into different task-specific feature spaces may benefit model learning, this paper designed a Task-aware Knowledge Injection module to differentiate the features in different task branches.

\subsubsection{Combining Graph and Transformer.}
Recently, the Transformer model has been introduced to deal with graph-structured data.
%
According to the relative position of the GNN and Transformer layers, current works could be divided into three architectures~\cite{min2022transformer}: 
(1) building Transformer blocks on top of GNN blocks;
(2) alternately stacking GNN and Transformer blocks~\shortcite{lin2021mesh};
and (3) parallelizing GNN and Transformer blocks~\shortcite{zhang2020graph}.
%
%
%
%
Most works~\cite{rong2020self, mialon2021graphit} adopted the first architecture. 
Especially, GraphTrans~\cite{wu2021representing} applied a permutation-invariant Transformer module after a standard GNN module to learn the high-level and long-range relationships.
Graph-Transformer architecture has also been introduced to handle the WSI analysis tasks~\cite{zheng2021deep}. 
However, the existing studies are limited to the single task setting and pay no attention to leveraging the domain knowledge from the pathologists for better model design.
\section{Methodology}

In this section, we elaborate on our designed multi-task framework for WSI analysis with specially designed Task-aware Knowledge Injection and Domain Knowledge-driven Graph Pooling modules.
Figure~\ref{GT_Framwork_Overview} shows the overview of the proposed framework. 
%
%
Given a WSI $X$, our framework predicts the labels of two tasks simultaneously: the slide-level tumor typing label $\hat{Y}_{type}$ and staging label $\hat{Y}_{stage}$.
%
%
%
Specifically, we first construct graph $\mathcal{G}$ followed by the task-shared Graph Convolution (GC) layer. 
After that, the framework is divided into two task-specific branches by the corresponding Task-specific Knowledge Injection modules. 
Further, the transferred task-specific graph representation is fed into the corresponding Domain Knowledge-driven Graph Pooling module for each branch. 
Finally, a sequence of task-specific Transformer layers followed by MLP are employed to predict slide-level labels of multiple tasks according to the pooled task-specific representations. 

%
%
%
%
%

\subsection{Graph-based Shared Feature Learning}
As illustrated in Figure~\ref{GT_Framwork_Overview}, given a WSI $X$ under $20\times$ magnification, we first apply the sliding window strategy to crop $X$ into numerous image tiles without overlap. 
%
%
Then we construct a graph $\mathcal{G} = \{\mathcal{V}, \mathcal{E}\}$, where $\mathcal{V}$ represents the extracted feature embeddings of the preserved image tiles. 
%
%
%
The edge set $\mathcal{E}$ represents the bordering relationships between image tiles in an 8-adjacent manner as shown in Figure~\ref{GT_Framwork_Overview}.
Then, the generated graph $\mathcal{G}$ is able to depict the feature and spatial relations of the WSI, and is thus suitable for further analysis.

Following the principle of a Graph-Transformer architecture, our framework first uses a Graph Convolution (GC) layer to extract the task-shared low-level representation of $\mathcal{G}$.
As illustrated in previous works~\cite{wu2021representing, NEURIPS2020_94aef384}, the GNN part in Graph-Transformer architecture learns the representation at graph nodes from neighborhood features. 
This neighborhood aggregation of GNN is helpful for learning local and short-range correlations of graph nodes, and thus suitable to serve as the shared layers for multiple different tasks.
%
%
The message propagation and aggregation of the graph are defined as
\begin{equation}
H_{{l}+1}=R e L U\left(\hat{A} H_{{l}} W_{{l}}\right), \quad l=1,2, \ldots, L.
\end{equation}
\begin{equation}
\hat{A}=\tilde{D}^{-\frac{1}{2}} \tilde{A} \tilde{D}^{-\frac{1}{2}},
\end{equation}
where $\tilde{A}=A+I_N$ is the adjacency matrix $A$ of graph $\mathcal{G}$ with added self-connections, and $I_N$ is the identity matrix. $\tilde{D}_{i i}=\sum_{j} \tilde{A}_{i j}$ is a diagonal matrix, and $W_{{l}}\in \mathbb{R}^{d\times d}$ is a layer-specific trainable weight matrix. $H_{{l}}\in \mathbb{R}^{|\mathcal{V}|\times d}$ is the input of the $l_{th}$ GC layer, where $|\mathcal{V}|$ is the number of nodes and $d$ is the dimension of each node, and  $H_{{l}}$ is initialized with the node features of $\mathcal{G}$.

\begin{figure}[t]
\centering
\includegraphics[width=0.9\linewidth]{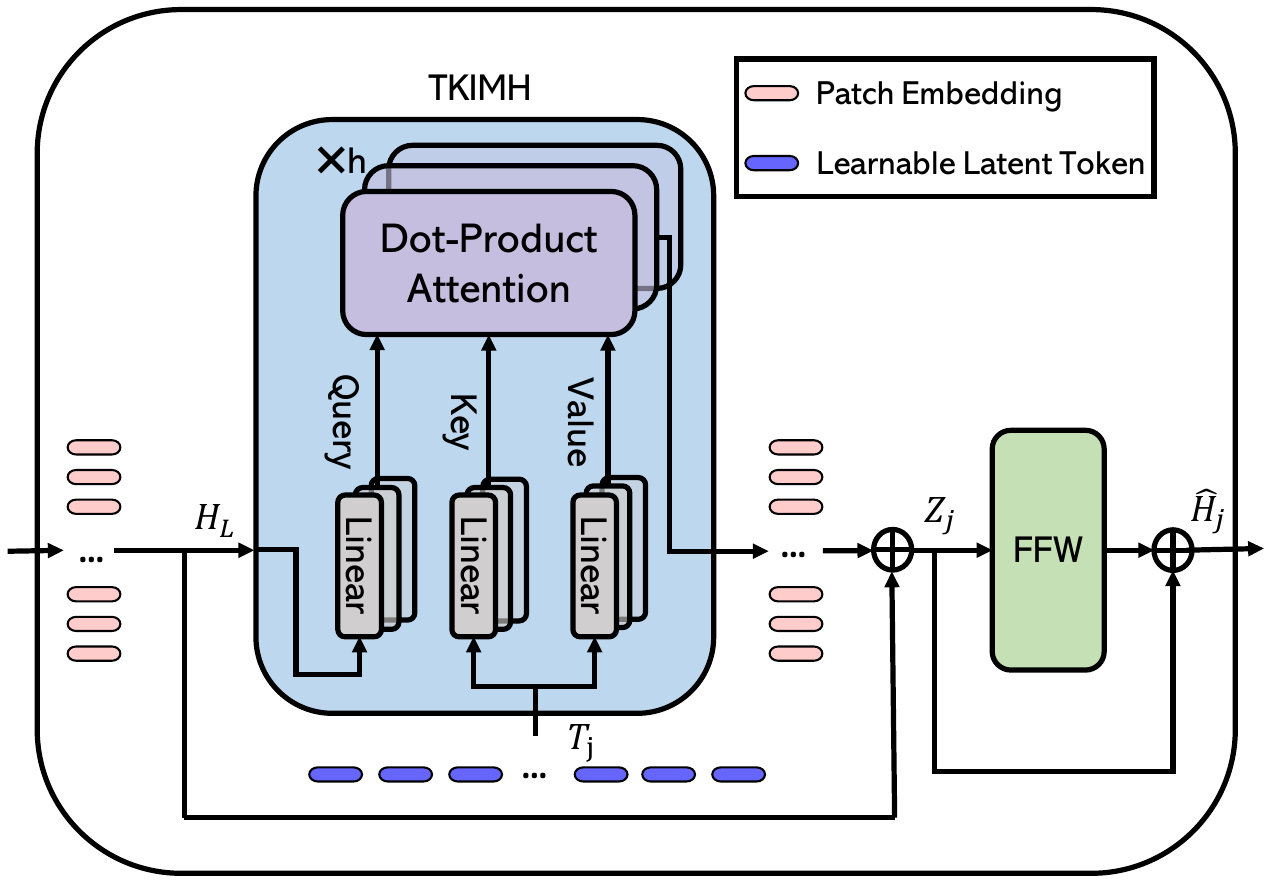}
\caption{Illustration of Task-aware Knowledge Injection.} 
\label{Task Knowledge Injection}
\end{figure}

\subsection{Task-aware Knowledge Injection}

For a more accurate representation learning for different tasks, we propose a Task-aware Knowledge Injection module to store the task-specific knowledge in different task branches and thus transfer the task-shared feature from the task-shared GCN into task-specific feature spaces.
%
%
%
The developed module calculates the correlation among the task-shared features with the task-specific knowledge based on the multi-head attention mechanism~\cite{vaswani2017attention}:
\begin{equation}
\operatorname{MH}(Q, K, V)=\left[O_{1}, \ldots, O_{h}\right] W^{O},
\end{equation}
\begin{equation}
O_{i}=\operatorname{Att}\left(Q W_{i}^{Q}, K W_{i}^{K}, V W_{i}^{V}\right),
\end{equation}
where $\operatorname{Att}(\boldsymbol{Q}, \boldsymbol{K}, \boldsymbol{V})=\sigma\left(\boldsymbol{Q} \boldsymbol{K}^{T}\right) \boldsymbol{V}$, $h$ is the number of parallel attention layers, and $\sigma$ is an activation function.
To transfer the task-shared features into task-specific spaces, we design a novel Task-aware Knowledge Injection multi-head cross attention (TKIMH) block, as shown in Figure~\ref{Task Knowledge Injection}.
Specifically, we take the task-shared hidden representation $H_{\mathrm{L}}$ as the query ($\boldsymbol{Q}$), and the task-specific trainable latent tokens $T_j$ as the key ($\boldsymbol{K}$) and value ($\boldsymbol{V}$) for the cross multi-head attention calculation. Each task branch has an independent set of trainable latent tokens, which is able to store the task-aware knowledge learned from the dataset during the training process.
The TKIMH for task $j$ can be denoted as:
\begin{equation}
\operatorname{TKIMH}(H_{\mathrm{L}}, T_j)=\left[O_{1j}, \ldots, O_{hj}\right] W^{O}_j,
\end{equation}
\begin{equation}
O_{ij}=\operatorname{Att}\left(H_{\mathrm{L}} W_{ij}^{Q}, T_j W_{ij}^{K}, T_j W_{ij}^{V}\right),
\end{equation}
where $H_{\mathrm{L}}\in \mathbb{R}^{|\mathcal{V}|\times d}$ is the task-shared hidden representation, $T_j\in \mathbb{R}^{m \times d}$ is the learnable latent tokens containing the task-specific knowledge for task $j$, $m$ is the number of the latent tokens,
$W_{ij}^{Q}, W_{ij}^{K}, W_{ij}^{V}\in \mathbb{R}^{d\times d}$ and $W^{O}_j\in \mathbb{R}^{hd\times d}$ are parameter matrices for linear projection operations for task $j$.
Using the ingredients above, the Task-aware Knowledge Injection module for task $j$ is defined as follows:
\begin{equation}
Z_j=\mathrm{LN}(H_{\mathrm{L}}+\operatorname{TKIMH}(H_{\mathrm{L}}, T_j)),
\end{equation}
\begin{equation}
\hat{H}_j=\mathrm{LN}(Z_j+\mathrm{rFF}(Z_j)),
\end{equation}
where $\mathrm{rFF}$ is a row-wise feedforward layer that processes each individual row independently and identically, LN is a layer normalization~\cite{ba2016layer}, and $\hat{H}_j$ is the transferred task-specific graph features for task $j$. 
Note that the above module can be easily adapted to any Graph-Transformer architecture.
%

%

\subsection{Domain Knowledge-driven Graph Pooling}
\begin{figure}[t]
\centering
\includegraphics[width=0.91\linewidth]{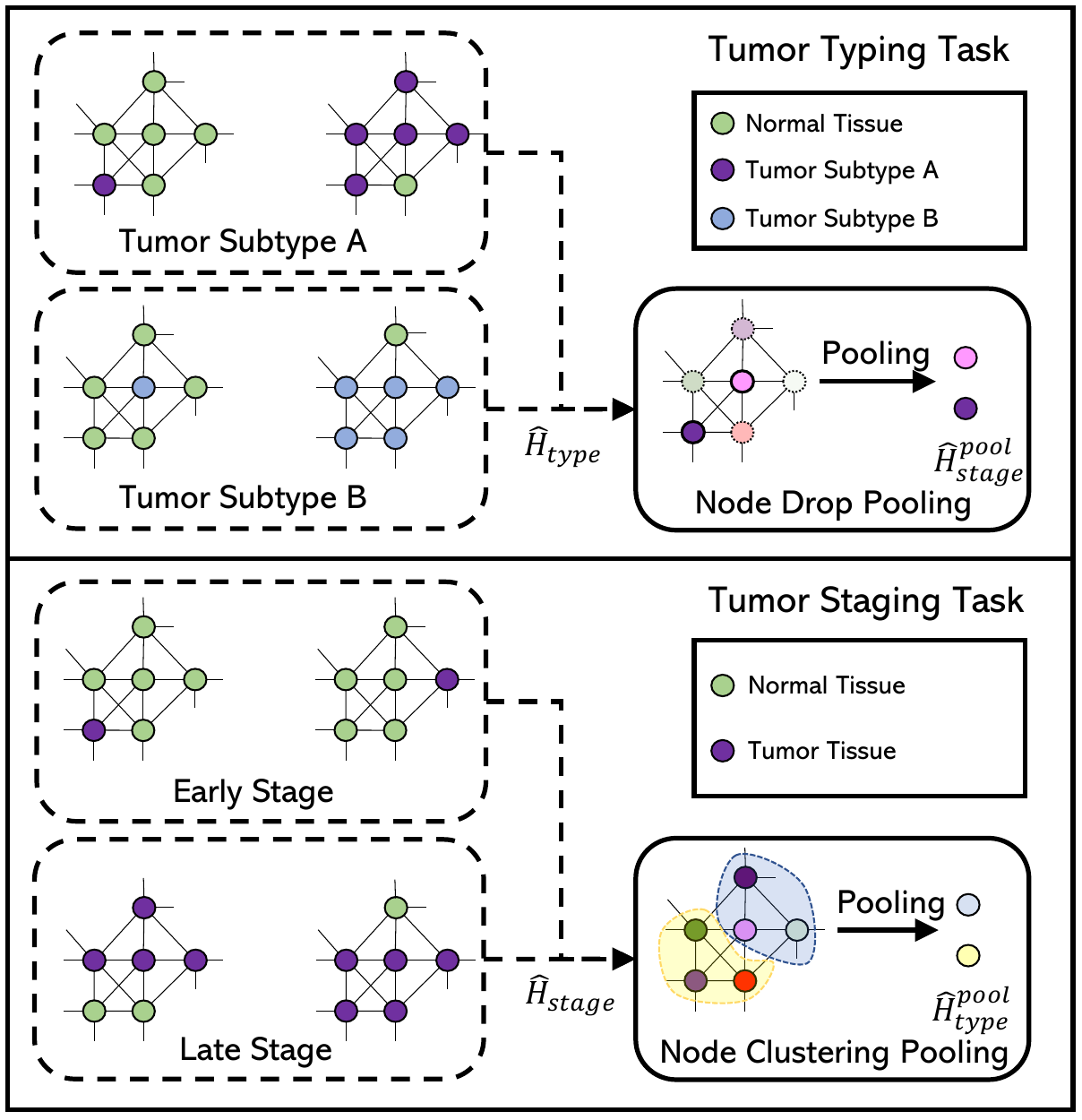}
\caption{Overview of Domain Knowledge-driven Pooling.} 
\label{Domain_Driven_Pooling}
\end{figure}

%
Domain Knowledge-driven Graph Pooling is developed by task-aware pooling methods to meet the requirements of different downstream tasks. 
%
%
%
%
%
%
As shown in Figure~\ref{Domain_Driven_Pooling}, we adopt two different graph pooling methods (\ie node drop method and node clustering method) for two tasks (\ie tumor staging and tumor typing) with different diagnosis patterns.
%

\paragraph{Node Drop Pooling for Typing.}
During the clinical diagnosis process,
the pathologists first examine the WSI to locate the tumor region and then determine the tumor type. 
%
Our node drop pooling method is designed to leverage the clinical process (as shown at the top of Figure~\ref{Domain_Driven_Pooling}). The model decision highly depends on the discriminative nodes (\ie, tumor subtype A/B node) instead of the ratio or shape of different types of nodes in the whole graph. 
%
%
Therefore, the node drop method will be sufficient for the tumor typing task as long as one of the tumor nodes can be preserved. 
As previous work~\cite{papp2021dropgnn} has also pointed out that random dropping will increase the expressiveness of GNN, we implemented a random and independent node dropping in each training runs to generate the task-aware pooled representation $\hat{H}_{type}^{pool}$ for tumor typing task. 
Compared with the rank-based dropping method, our scheme will make the task more challenging and serve as a data augmentation method, which will make the corresponding branch more robust and more powerful in detecting discriminative image patches.

\paragraph{Node Clustering Pooling for Staging.}
Several elements influence the tumor stage diagnosis results of pathologists, including abnormal cells, the presence and size of tumor regions, and metastatic tumors.
%
In general, the ratio and the shape of the tumor tissue nodes in the whole graph will be essential for the slide-level tumor stage diagnosis, as observed in the bottom of Figure~\ref{Domain_Driven_Pooling}. 
%
%
Node clustering pooling methods are more suitable for the staging task to preserve the whole graph information, as node drop methods may lose the above information during dropping.
%
%
%
Inspired by GMPool~\shortcite{baek2021accurate}, We design GCMinCut, an improved version of MinCut Pooling~\cite{bianchi2020spectral},
in which we replaced the MLP during the pooling with an additional GC layer to import the neighboring information of the graph.
The GCMinCut pooling can be denoted as follows:
\begin{equation}
\mathbf{S}=R e L U\left(\hat{A} \hat{H}_{stage} W_{\mathrm{pool}}\right),
\hat{H}_{stage}^{\text {pool }}=\mathbf{S}^{T} \hat{H}_{stage},
\end{equation}
where $\hat{H}_{stage}\in \mathbb{R}^{|\mathcal{V}|\times d}$ is the task-specific representation transferred by the task-aware knowledge injection module in the tumor staging branch, $W_{\mathrm{pool}}\in \mathbb{R}^{d\times p}$ is a trainable weight matrix, and $\mathbf{S} \in \mathbb{R}^{ |\mathcal{V}| \times p}$ is the assignment matrix for soft node clustering, $p$ is the number of node clusters (\ie, the number of nodes after pooling), and $\hat{H}_{stage}^{pool}\in \mathbb{R}^{p\times d}$ is the task-aware pooled representation for tumor staging task.

\subsection{Technical Details and Training Procedure}
%
%
%
After the Domain Knowledge-driven Graph Pooling module, the task-aware pooled representations are fed into a standard Transformer layer stack with no additive positional embeddings as the GNN has already encoded the structural information into the node embeddings.
After that, we apply task-specific MLP heads for each branch to predict the task labels. 
The label prediction $\hat{Y}_{i}$ of task $i$ can be denoted as:
\begin{equation}
\hat{X}_{i}= \operatorname{Transformer}\left( [CLS; \hat{H}_{i}^{\text {pool }}] \right),
\hat{Y}_{i}=\operatorname{MLP}\left(\hat{X}_{i}^{(0)} \right),
\end{equation}
where $CLS\in \mathbb{R}^{1\times d}$ is the class token in Transformer.

To train the network, we first employed the cross-entropy loss for both tasks. 
Take the type prediction task as an example, the objectiveness is:
\begin{equation}
\mathcal{L}_{type}=-\frac{1}{N} \sum_{i=1}^{N} \sum_{j=1}^{C_{type}} Y_{type}^{(ij)} \log \left(\hat{Y}_{type}^{(ij)}\right),
\end{equation}
%
%
where $N$ is the whole sample number, $C_{type}$ is category number for type prediction task, and $Y$ is the one-hot label.

%
Then unsupervised MinCut pooling loss~\cite{bianchi2020spectral} is adopted for extra regularization:
\begin{equation}
\mathcal{L}_{mincut}=\underbrace{-\frac{\operatorname{Tr}\left(\mathbf{S}^{T} \tilde{\mathbf{A}} \mathbf{S}\right)}{\operatorname{Tr}\left(\mathbf{S}^{T} \tilde{\mathbf{D}} \mathbf{S}\right)}}_{\mathcal{L}_{c}}+\underbrace{\left\|\frac{\mathbf{S}^{T} \mathbf{S}}{\left\|\mathbf{S}^{T} \mathbf{S}\right\|_{F}}-\frac{\mathbf{I}_{p}}{\sqrt{p}}\right\|_{F}}_{\mathcal{L}_{o}},
\end{equation}
where $\|\cdot\| F$ indicates the Frobenius norm.
%
The cut loss term $\mathcal{L}_c$ encourages strongly connected nodes to be clustered together, and the orthogonality loss term $\mathcal{L}_o$ encourages the cluster assignments to be a similar size.
Finally, the total loss $\mathcal{L}_{total}$ can be denoted as the weighted summation of the above losses:
\begin{equation}
\mathcal{L}_{total}=w_{t}\mathcal{L}_{type}+w_{s}\mathcal{L}_{stage}+w_{m}\mathcal{L}_{mincut}.
\end{equation}
\section{Experiments}

\begin{table*}[ht]
\small
\centering
\resizebox{0.85\textwidth}{!}{%
\begin{tabular}{lccc|ccc}
\toprule[1pt]
\multirow{2}{*}{{Method}} &
\multicolumn{3}{c}{{Typing}} & \multicolumn{3}{c}{{Staging}} \\ 
\cmidrule(r){2-4}\cmidrule(r){5-7} & {AUC} & {ACC} & {F1} & {AUC} & {ACC} & {F1} \\ 
\bottomrule[1pt]

ABMIL~\shortcite{ilse2018attention} &
$95.42\pm2.02$ & $89.90\pm2.77$ & $89.82\pm2.75$ & $75.35\pm3.74$ & $70.51\pm1.88$ & $58.54\pm2.55$ \\
Gated-ABMIL~\shortcite{ilse2018attention} &
$94.84\pm1.60$ & $88.63\pm2.98$ & $88.61\pm2.87$ & $73.65\pm3.25$ & $69.69\pm2.33$ & $58.65\pm2.78$ \\
DeepAttnMIL~\shortcite{yao2020whole} &
$96.87\pm1.44$ & $91.37\pm2.53$ & $91.37\pm2.49$ & $76.53\pm2.84$ & $70.32\pm2.19$ & $58.44\pm2.86$ \\
CLAM-MIL~\shortcite{lu2021data} &
$84.93\pm3.15$ & $79.46\pm2.91$ & $78.57\pm3.27$ & $70.97\pm3.20$ & $70.32\pm2.20$ & $58.64\pm3.17$ \\
CLAM-SB~\shortcite{lu2021data} &
$95.69\pm2.31$ & $90.62\pm2.93$ & $90.60\pm2.96$ & $74.94\pm4.22$ & $70.39\pm2.22$ & $58.28\pm2.78$ \\
DS-MIL~\shortcite{li2021dual} &
$93.97\pm2.52$ & $87.08\pm3.04$ & $86.90\pm3.12$ & $73.21\pm4.35$ & $68.94\pm2.37$ & $59.31\pm2.31$ \\ 
GT-MIL~\shortcite{zheng2021deep} &
$97.20\pm1.19$ & $92.31\pm2.52$ & $92.33\pm2.46$ & $78.63\pm3.56$ & $71.20\pm3.60$ & $68.38\pm3.37$ \\ 
Trans-MIL~\shortcite{wang2022lymph} &
$95.56\pm2.11$ & $89.14\pm3.30$ & $89.04\pm3.31$ & $73.34\pm3.15$ & $68.56\pm3.46$ & $57.70\pm2.34$ \\ 
\bottomrule[1pt]

\textbf{Ours} &
$\textbf{98.44}\pm\textbf{0.67}$ & $\textbf{93.89}\pm\textbf{1.60}$ & $\textbf{93.89}\pm\textbf{1.59}$ & $\textbf{80.22}\pm\textbf{1.94}$ & $\textbf{74.98}\pm\textbf{3.08}$ & $\textbf{72.55}\pm\textbf{2.48}$  \\
\bottomrule[1pt]
\end{tabular}
}
\caption{Comparison with other methods on KICA dataset. Top results are shown in bold.
}
\label{Comparison with Single-Task SOTAs on KICA Dataset}
\end{table*}

\begin{table*}[ht]
\small
\centering
\resizebox{0.85\textwidth}{!}{%
\begin{tabular}{lccc|ccc}
\toprule[1pt]
\multirow{2}{*}{{Method}} &
\multicolumn{3}{c}{{Typing}} & \multicolumn{3}{c}{{Staging}} \\ 
\cmidrule(r){2-4}\cmidrule(r){5-7} & {AUC} & {ACC} & {F1} & {AUC} & {ACC} & {F1} \\ 
\bottomrule[1pt]

ABMIL~\shortcite{ilse2018attention} &
$92.51\pm3.39$ & $86.47\pm4.16$ & $86.33\pm4.23$ & 
$53.01\pm3.95$ & $54.34\pm3.02$ & $51.36\pm3.19$ \\ 
Gated-ABMIL~\shortcite{ilse2018attention} &
$95.17\pm2.47$ & $88.54\pm3.05$ & $88.39\pm3.12$ & 
$50.64\pm4.12$ & $53.38\pm4.22$ & $53.54\pm5.02$ \\ 
DeepAttnMIL~\shortcite{yao2020whole} &
$96.12\pm1.84$ & $90.64\pm2.93$ & $90.50\pm3.07$ & 
$61.87\pm3.32$ & $61.48\pm4.28$ & $50.59\pm2.79$ \\ 
CLAM-MIL~\shortcite{lu2021data} &
$77.89\pm5.90$ & $73.98\pm5.25$ & $73.55\pm5.57$ & 
$61.23\pm4.15$ & $58.38\pm4.11$ & $50.38\pm4.14$ \\ 
CLAM-SB~\shortcite{lu2021data} &
$95.85\pm1.78$ & $90.66\pm3.02$ & $90.57\pm3.10$ & 
$63.01\pm3.05$ & $59.96\pm3.46$ & $51.75\pm3.79$ \\ 
DS-MIL~\shortcite{li2021dual} &
$87.80\pm3.97$ & $81.63\pm4.81$ & $81.05\pm5.16$ & 
$61.75\pm2.20$ & $59.39\pm3.30$ & $54.36\pm5.27$ \\ 
GT-MIL~\shortcite{zheng2021deep} &
$95.93\pm1.58$ & $89.87\pm3.64$ & $89.83\pm3.60$ & 
$69.23\pm3.64$ & $65.20\pm3.72$ & $62.64\pm3.22$ \\ 
Trans-MIL~\shortcite{wang2022lymph} &
$94.24\pm2.33$ & $86.59\pm3.17$ & $86.48\pm3.14$ & 
$60.56\pm4.72$ & $61.47\pm3.87$ & $49.73\pm3.32$ \\ 
\bottomrule[1pt]

\textbf{Ours} &
$\textbf{97.49}\pm\textbf{1.46}$ & $\textbf{92.81}\pm\textbf{2.35}$ & $\textbf{92.74}\pm\textbf{2.41}$ & $\textbf{71.48}\pm\textbf{3.42}$ & $\textbf{66.63}\pm\textbf{3.14}$ & $\textbf{65.73}\pm\textbf{2.83}$  \\
\bottomrule[1pt]
\end{tabular}
}
\caption{Comparison with other methods on ESCA dataset. 
Top results are shown in bold.
}
\label{Comparison with Single-Task SOTAs on ESCA Dataset}
\end{table*}
\begin{table*}[ht]
\small
\centering
\resizebox{0.9\textwidth}{!}{%
\begin{tabular}{ccccc|ccc}
\toprule[1pt]
\multirow{2}{*}{DomainPool} &\multirow{2}{*}{TK-Injection} &
\multicolumn{3}{c}{Typing} & \multicolumn{3}{c}{Staging} \\ 
\cmidrule(r){3-5}\cmidrule(r){6-8} & & AUC & ACC & F1 & AUC & ACC & F1 \\ 
\bottomrule[1pt]
Drop-based  &  &
$95.72\pm1.64$ & $90.11\pm2.39$ & $90.01\pm2.35$ & $77.07\pm2.33$ & $71.78\pm2.78$ & $69.93\pm3.70$ \\ 
Cluster-based &  &
$97.40\pm0.99$ & $91.97\pm2.27$ & $92.02\pm2.13$ & $80.12\pm3.51$ & $73.45\pm2.82$ & $71.47\pm3.68$ \\ 
\checkmark  &  &
$97.90\pm1.32$ & $93.50\pm1.91$ & $93.53\pm1.88$ & $\textbf{80.67}\pm\textbf{3.51}$ & $74.07\pm3.53$ & $70.78\pm3.33$ \\ 
\checkmark  & Linear &
$98.10\pm0.55$ & $93.59\pm1.56$ & $93.55\pm1.61$ & 
$79.86\pm2.20$ & $73.57\pm2.94$ & $71.06\pm3.13$ \\ 
\checkmark & \checkmark&
$\textbf{98.44}\pm\textbf{0.67}$ & $\textbf{93.89}\pm\textbf{1.60}$ & $\textbf{93.89}\pm\textbf{1.59}$ & $80.22\pm1.94$ & $\textbf{74.98}\pm\textbf{3.08}$ & $\textbf{72.55}\pm\textbf{2.48}$ \\
\bottomrule[1pt]
\end{tabular}
}
\caption{Ablation study on KICA dataset. DomainPool and TK-Injection denote the Domain Knowledge-driven Graph Pooling module and the Task-aware Knowledge Injection module, respectively. Drop-based and Cluster-based denote that replacing the DomainPool with node drop pooling methods or node clustering pooling methods, respectively.
Linear denotes that replacing the cross-attention mechanism in the Task-aware Knowledge Injection module with task-specific linear projections.}
\label{Ablation Study on KICA Dataset}
\end{table*}
%

%

\begin{table}[ht]
\small
\centering
\begin{tabular}{cccc|ccc}
\toprule[1pt]
\multirow{2}{*}{Paradigm} &
\multicolumn{3}{c}{Typing} & \multicolumn{3}{c}{Staging} \\ 
\cmidrule(r){2-4}\cmidrule(r){5-7} & AUC & ACC &F1 & AUC & ACC & F1\\ 
\bottomrule[1pt]
Single&
$98.41$ & $93.07$ & $93.08$ & $80.14$ & $73.67$ & $71.45$\\ 
Multi&
$\textbf{98.44}$ & $\textbf{93.89}$ & $\textbf{93.89}$ & 
$\textbf{80.22}$ & $\textbf{74.98}$ & $\textbf{72.55}$\\ 
\bottomrule[1pt]
\end{tabular}
\caption{Comparison of single-task and multi-task paradigm on MulGT. Single for single-task while Multi for multi-task.}
\label{Effectiveness of Multi-task Learning Paradigm}
\end{table}

\begin{table}[ht]
\small
\centering
\begin{tabular}{cccc|ccc}
\toprule[1pt]
\multirow{2}{*}{Scheme} &
\multicolumn{3}{c}{Typing} & \multicolumn{3}{c}{Staging} \\ 
\cmidrule(r){2-4}\cmidrule(r){5-7} & AUC & ACC & F1 & AUC & ACC & F1\\ 
\bottomrule[1pt]
Shared&
$98.23$ & $93.30$ & $92.32$ & $78.54$ & $72.71$ & $69.28$\\ 
\textbf{Specific}&
$\textbf{98.44}$ & $\textbf{93.89}$ & $\textbf{93.89}$ & 
$\textbf{80.22}$ & $\textbf{74.98}$ & $\textbf{72.55}$\\ 
\bottomrule[1pt]
\end{tabular}
\caption{Comparison of different task-knowledge latent token schemes in MulGT. Shared means using a shared set of latent tokens, while Specific means using independent sets of latent tokens in different task branches.}
\label{Comparison of Separated and Shared Latent Tokens in Task Knowledge Injection Module}
\end{table}

\subsection{Datasets}
We evaluate the proposed MulGT framework
on two public datasets (KICA and ESCA) from The Cancer Genome Atlas (TCGA) repository. 
In the tumor staging task, patients with TNM labels as I/II stage are categorized as the early stage while patients with TNM labels as III/IV are categorized as the late stage. 
We excluded patients with missing diagnostic WSI, tumor type diagnosis, and TNM label. 
The details of the above two datasets are as follows:
\begin{itemize}
    \item \textbf{KICA} is the kidney carcinoma project containing 371 cases with 279 early-stage cases and 92 late-stage cases. 
    %
    For the tumor typing task, there are 259 cases diagnosed as kidney renal papillary cell carcinoma and 112 cases diagnosed as kidney chromophobe.
%
    %
    %
    \item \textbf{ESCA} is the esophageal carcinoma project containing 161 cases with 96 early-stage cases and 65 late-stage cases. 
    %
    %
    For the tumor typing task, there are 94 cases diagnosed as adenomas and adenocaricinomas and 67 cases diagnosed as squamous cell carcinoma.
\end{itemize}

\subsection{Experimental Setup}
The proposed framework was implemented using PyTorch~\cite{paszke2019pytorch} and PyTorch Geometric~\cite{fey2019fast} frameworks. All experiments were conducted on a workstation with four NVIDIA RTX 3090 GPUs. 
For a fair comparison, the proposed framework and other SOTA methods were all tested using non-overlapping $512 \times 512$ image tiles cropped under $20 \times$ magnification from the WSIs, we filtered out image tiles containing less than 85\% tissue region. Besides, KimiaNet~\cite{riasatian2020kimianet} served as the feature extractor for all methods to convert each $512 \times 512$ image tile into 1024-dimensional features for graph initialization. 
We set the number of nodes after graph pooling as 100, following the setting in GT-MIL~\shortcite{zheng2021deep}.
We select the number of latent tokens of Task-aware Knowledge Injection module as $150$ by hyper-parameter searching.
All methods are trained with a batch size of 8 for 40 epochs with the Adam optimizer.
%
For evaluation, the area under the curve (AUC) of receiver operating characteristic, the accuracy (ACC), and the F1-score were adopted. 
All approaches were evaluated with five-fold cross-validation from three different runs (initializations). 
%
%
\subsection{Experimental Results}
\subsubsection{Comparison with State-of-the-art Methods.}
We compare our framework with eight single-task state-of-the-art (SOTA) methods for WSI analysis including: 
(1) ABMIL~\shortcite{ilse2018attention}, 
(2) Gated-ABMIL~\shortcite{ilse2018attention}, 
(3) CLAM-MIL~\shortcite{lu2021data}, 
(4) CLAM-SB~\shortcite{lu2021data}, 
(5) DeepAttnMIL~\shortcite{yao2020whole}, 
(6) DS-MIL~\shortcite{li2021dual}, 
(7) GT-MIL~\shortcite{zheng2021deep}, and 
(8) Trans-MIL~\shortcite{wang2022lymph}. 
For DS-MIL~\shortcite{li2021dual} method, it was introduced with a pyramidal fusion mechanism for multi-scale WSI features. We only test its performance under the single-scale setting for a fair comparison. 
The results for KICA and ESCA datasets are summarized in Table~\ref{Comparison with Single-Task SOTAs on KICA Dataset} and Table~\ref{Comparison with Single-Task SOTAs on ESCA Dataset}, respectively.
Overall, across all tasks and different datasets, our frameworks consistently achieve the highest performance on all the evaluation metrics. 
GT-MIL~\shortcite{zheng2021deep} performs best among the previous SOTAs, which demonstrates the powerful representation of Graph-Transformer architecture in WSI analysis. 
However, compared with our methods, GT-MIL~\shortcite{zheng2021deep} only built a Graph-Transformer network in the single-task setting and only adopted MinCut pooling~\cite{bianchi2020spectral}.
In comparison with GT-MIL~\shortcite{zheng2021deep}, for instance, our framework achieved a performance increase of $\textbf{1.24\%}$ in AUC, $\textbf{1.58\%}$ in ACC, $\textbf{1.56\%}$ in F1-score for tumor typing task, and $\textbf{1.59\%}$ in AUC, $\textbf{3.78\%}$ in ACC, $\textbf{4.17\%}$ in F1-score for tumor staging task on KICA dataset, which validates the effectiveness of the designs in our framework.
More importantly, we observe an obvious improvement in tumor staging tasks, which is more challenging among the two tasks. 
The reason is probably the more general and robust task-shared representations learned in the multi-task learning paradigm.
\subsubsection{Ablation Study.}
We conducted an ablation study on KICA dataset to demonstrate the effectiveness of each proposed component. 
%
We first compare our Domain Knowledge-driven Graph Pooling module with node drop based methods as well as node clustering based methods. 
We test multiple node drop based methods (SortPool~\shortcite{zhang2018end}, TopKPool~\shortcite{gao2019graph}, and SAGPool~\shortcite{lee2019self}) and node clustering based methods (DiffPool~\shortcite{ying2018hierarchical}, MinCutPool~\shortcite{bianchi2020spectral}, and GMPool~\shortcite{baek2021accurate}), and report the best performance in the above two groups. As observed from the first three rows in Table~\ref{Ablation Study on KICA Dataset},
%
obvious improvement could be seen in all evaluation metrics except F1 in tumor staging, which demonstrates the effectiveness of exploiting the domain knowledge
for different tasks
during the graph pooling process.
The effectiveness of the proposed TK-Injection module is shown by comparison with baselines with no task-specific transferring and simple task-specific linear projections from the last three rows in Table~\ref{Ablation Study on KICA Dataset}.
Compared with baselines without task-specific transferring, performance increases can be found in all the evaluation metrics except AUC and ACC in tumor staging in task-specific linear projections, which demonstrate that it is essential to transfer the task-agnostic feature into different task-specific spaces during multi-task learning. 
Moreover, our cross-attention based task-ware knowledge injection module performs better in all aspects than task-specific linear projections, which illustrate the effectiveness of storing the task-specific knowledge in latent tokens and importing them via the attention mechanism.

\subsubsection{Investigation of Multi-task Learning Paradigm.}
To figure out the effectiveness of the multi-task learning paradigm, we also tested our framework and the elaborately designed modules under the single-task setting on KICA dataset. 
%
Table~\ref{Effectiveness of Multi-task Learning Paradigm} summarizes the experimental results, where the multi-task paradigm benefited both tasks, especially the more challenging one, \ie, tumor staging. 
The ACC and F1-score in tumor staging task increase by $1.31\%$ and $1.10\%$, respectively. 
The performance improvement in tumor typing is less than staging, as it has already achieved very high performance.
%
However, note that our framework still outperforms all previous SOTAs in Table~\ref{Comparison with Single-Task SOTAs on KICA Dataset} including the GT-MIL~\shortcite{zheng2021deep} under the single-task setting, which means that our Task-aware Knowledge Injection and Domain Knowledge-driven Pooling modules could also improve the performance of single-task based methods.

\subsubsection{Investigation of Task-knowledge Latent Token.}
We also investigate the impact of different schemes for task-aware knowledge latent tokens on KICA dataset, and show the mean results in Table~\ref{Comparison of Separated and Shared Latent Tokens in Task Knowledge Injection Module}. 
The ``Shared" scheme means that different task branches use a shared set of knowledge latent tokens, while the ``Specific" scheme means that different task branches have independent knowledge latent token sets. 
The ``Specific" scheme achieves better performance in all metrics, especially in F1-score for typing task and ACC and F1-score in staging task. 
These experimental results show that different diagnostic tasks require different knowledge (sets), which is the same as the pathologists' experiences.
%
%

%
\begin{figure}[t]
\centering
\includegraphics[width=1.0\linewidth]{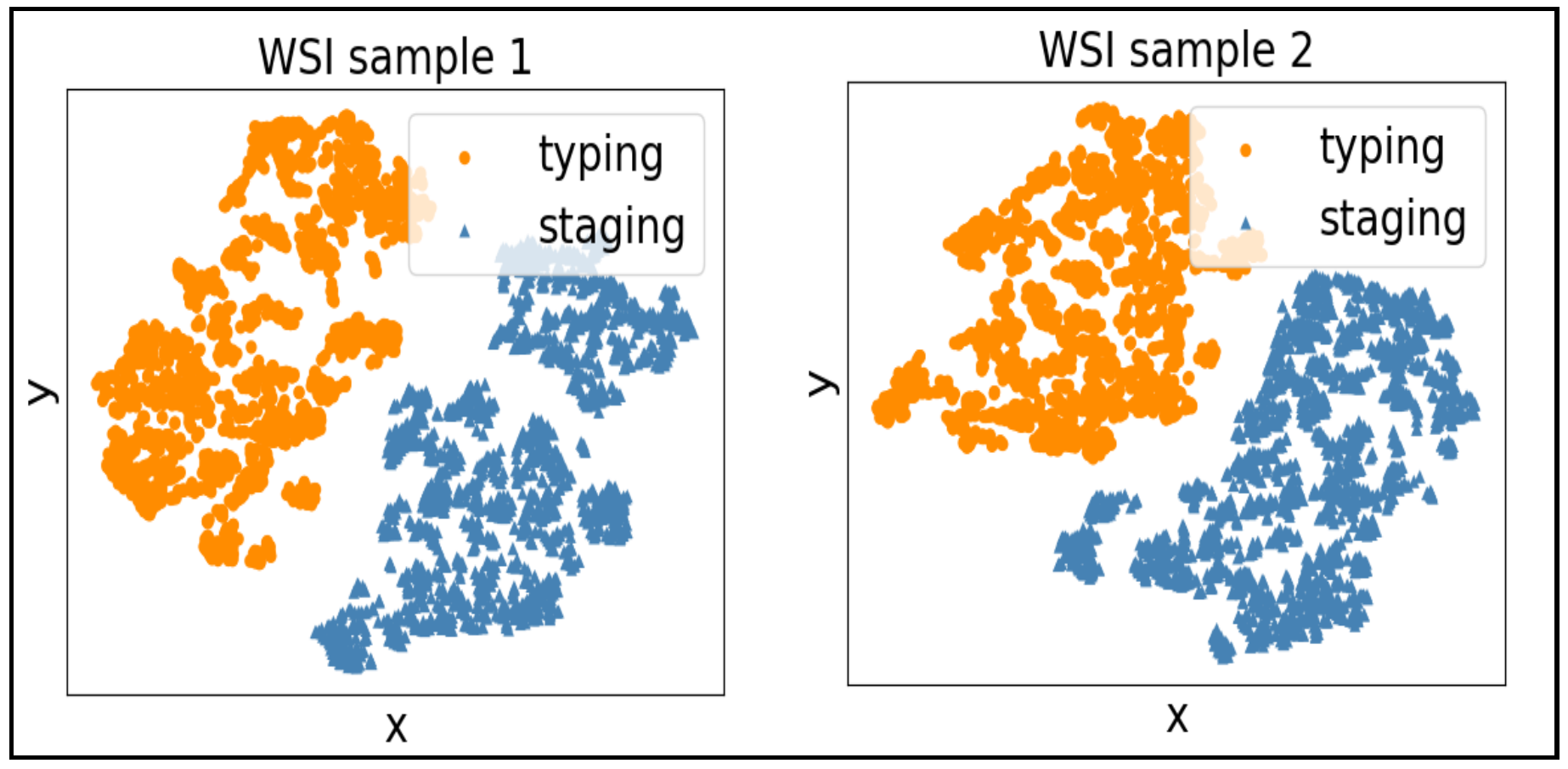}
\caption{t-SNE visualization of task-specific features in different Task-aware Knowledge Injection modules.
%
} 
\label{tsne_vis}
\end{figure}
\subsubsection{Visualization of Task-specific Feature Spaces.}
%
We further conduct t-SNE visualization of two different WSIs to demonstrate the learned task-specific features in Figure~\ref{tsne_vis}. 
The blue and orange nodes denote transferred node features in the Task-aware Knowledge Injection modules of typing branch and staging branch, respectively.
All samples show a clear separation of the nodes with different colors, which means the task-shared features are indeed successfully transferred into different task-specific feature spaces.
\section{Conclusion}
In this paper, we propose a novel MulGT framework for WSI analysis with multi-task learning. 
By exploring the commonalities and different diagnosis patterns in different WSI diagnosis tasks, our framework is able to learn more general and robust task-shared representation as well as more accurate task-specific features. 
Specially, a Task-aware Knowledge Injection module is introduced to store and import the knowledge of different tasks, thus transferring the task-shared representation into different task-specific feature spaces. 
Meanwhile, to leverage the domain knowledge from the pathologists, a Domain Knowledge-driven Graph Pooling module is elaborately designed to simulate the diagnosis pattern of different analysis tasks. Above building commons lead to performance improvement on both tasks.
Extensive experiments validate the prominence of the proposed framework.
In the future, we will extend our framework to other WSI analysis tasks, such as survival prediction and prognosis analysis, with domain knowledge from pathologists.
Meanwhile, we will develop a hierarchical multi-task Graph-Transformer framework to leverage the natural image pyramid structure of WSI for multi-scale analysis.
%

\section*{Acknowledgements} The work described in this paper was supported in part by
a grant from the Research Grants Council of the Hong Kong SAR, China (Project No. T45-401/22-N) and in part by HKU Seed Fund for Basic Research (Project No. 202009185079 and 202111159073).
The computations in this paper were partly performed using research computing facilities offered by Information Technology Services, The University of Hong Kong.

\bibliography{reference}

\end{document}